\documentclass[conference]{IEEEtran}
\IEEEoverridecommandlockouts
\usepackage[T1]{fontenc}

\usepackage{amsmath}
\usepackage{amssymb}
\usepackage{bm}          

\usepackage{graphicx}
\usepackage[caption=false,font=footnotesize]{subfig} 

\usepackage{booktabs}    
\usepackage{multirow}

\usepackage{algorithm}
\usepackage{algorithmic} 

\usepackage{cite}

\usepackage{xcolor}      
\usepackage{microtype}   

\usepackage{tabularx}

\usepackage[hidelinks]{hyperref}
\usepackage{orcidlink}

\begin{document}
\title{Calibrated Uncertainty for Informative Path Planning
in Aquatic Environmental Monitoring}
\author{
    \IEEEauthorblockN{Samuel Yanes Luis\IEEEauthorrefmark{1}\,\orcidlink{0000-0002-7796-3599},
    Alejandro Casado Pérez\IEEEauthorrefmark{2}\,\orcidlink{0009-0007-5572-5913},
    Alejandro Mendoza Barrionuevo\IEEEauthorrefmark{2}\,\orcidlink{0000-0002-6931-7124},\\
    Dame Seck Diop\IEEEauthorrefmark{2}\,\orcidlink{0009-0004-0752-1041},
    Sergio Toral Marín\IEEEauthorrefmark{2}\,\orcidlink{0000-0003-2612-0388} and
    Daniel Gutiérrez Reina\IEEEauthorrefmark{2}\,\orcidlink{0000-0002-9017-7349}}
    \IEEEauthorblockA{\IEEEauthorrefmark{1}Department of Electronic Technology, University of Sevilla, Sevilla, Spain}
    \IEEEauthorblockA{\IEEEauthorrefmark{2}Department of Electronic Engineering, University of Sevilla, Sevilla, Spain}
    \thanks{Corresponding author: S. Yanes Luis (e-mail: syanes@us.es). Other authors' e-mails:
    \{acasado4, amendoza1, dseck, storal, dgutierrezreina\}@us.es.}
}

\maketitle              
\begin{abstract}
Informative Path Planning for scalar field reconstruction uses predictive
uncertainty to direct sensing vehicles toward maximally informative locations.
Gaussian Processes provide this signal but their stationary isotropic kernels
are misspecified for non-homogeneous phenomena such as oil spills, producing
miscalibrated estimates that degrade planning. We investigate whether replacing
the Gaussian Process with a well-calibrated Deep Ensemble improves path planning
outcomes, and whether uncertainty quality interacts with the choice of planning
algorithm. Five strategies ($\epsilon$-Greedy, Value Greedy, Uncertainty Greedy,
Monte Carlo Tree Search, and Receding Horizon Orienteering) share a common Deep
Ensemble backbone trained on physics-based oil spill simulations. On held-out
stochastic spill scenarios, the Deep Ensemble reduces normalised reconstruction
error by $83\%$ relative to the Gaussian Process baseline. Crucially,
well-calibrated uncertainty amplifies the importance of the planning strategy:
the performance gap between algorithms is negligible under miscalibrated models
but becomes substantial under the ensemble, where multi-step lookahead planners
outperform greedy selection by up to $32\%$ in reconstruction error and achieve
IoU above $0.85$. Monte Carlo Tree Search is the recommended planner, matching
Orienteering in reconstruction quality at an order-of-magnitude lower
computational cost.
\end{abstract}


\section{Introduction}

Autonomous surface vehicles (ASVs) have emerged as a practical platform for in-situ
environmental monitoring of aquatic phenomena such as hydrocarbon spills, algal blooms,
and chemical plumes~\cite{mansfield2024survey}. In these scenarios, the vehicle must
reconstruct an unknown scalar field $f: \mathcal{X} \subset \mathbb{R}^2 \rightarrow \mathbb{R}$
from a limited budget of sequential onboard observations, where the order and location of
measurements critically determines the quality of the final reconstruction. This is
naturally framed as an active learning problem: rather than collecting data along a
pre-defined lawnmower path, the vehicle should allocate its measurement budget to
locations that are maximally informative~\cite{chen2019robotic}.

The dominant criterion for guiding this process is predictive uncertainty. When a model
provides uncertainty estimates that reliably correlate with reconstruction error,
navigating toward regions of maximum uncertainty is equivalent to navigating toward
regions of maximum expected improvement in reconstruction quality. This principle
underlies Informative Path Planning (IPP) with Gaussian Processes (GPs), where the
posterior variance is available in closed form and has historically been treated as a
tractable surrogate for the true error surface~\cite{yanes2023censored,samaniego2021bayesian}.
However, this surrogate is only reliable when the model assumptions match the phenomenon
under study. Classical GP formulations rely on stationary, isotropic kernels such as the
RBF, which impose spatially uniform smoothness through a single global length-scale.
Oil spill fields violate this assumption fundamentally, as sharp gradients at contamination
boundaries coexist with large smooth background regions, producing uncertainty estimates
that are systematically misaligned with the actual reconstruction error~\cite{yanes2024deep}.

This paper investigates whether replacing the GP with a deep learning model that provides
well-calibrated uncertainty can improve path planning outcomes. We build on recent work
showing that an ensemble of U-Net architectures trained on physics-based oil spill
simulations achieves significantly lower reconstruction error and better uncertainty
calibration than GP baselines~\cite{yanes2026uncertainty}. Well-calibrated epistemic
uncertainty constitutes the best available criterion for information-driven path selection,
since the vehicle need not access ground truth to assess where its estimate is least
reliable. Moreover, scalar field reconstruction under an uncertainty-based objective is
submodular, which motivates greedy algorithms with well-known approximation
guarantees~\cite{nemhauser1978analysis}, while also opening the door to more expressive
planners that trade computational cost for improved path efficiency. We implement and
evaluate a portfolio of planning strategies ranging from greedy selection to Monte Carlo
Tree Search (MCTS) and orienteering with simulated annealing, all sharing the same deep
ensemble model and evaluated on a common set of physics-based oil spill scenarios to
provide a controlled comparison of planning quality as a function of algorithm choice.
The remainder of the paper is organized as follows. Section~\ref{sec:related} reviews
related work.
Section~\ref{sec:methods} describes the uncertainty model and planning algorithms.
Section~\ref{sec:results} presents results, and Section~\ref{sec:conclusion} concludes.

\section{Related Work}
\label{sec:related}

Gaussian Processes have been the standard tool for scalar field reconstruction
in autonomous monitoring missions due to their native probabilistic formulation:
the posterior mean provides a point estimate of the field, while the posterior
variance is available in closed form and well-calibrated by construction under
the model assumptions~\cite{rasmussen2006gaussian}. Their application to aquatic
monitoring with ASVs has been extensively studied~\cite{samaniego2021bayesian,
yanes2024deep}, and sparse GP variants partially address the cubic scaling
bottleneck~\cite{titsias2009variational}. However, the stationary kernel
assumption remains a fundamental limitation: an RBF kernel enforces a single
global length-scale that is misspecified for phenomena such as oil spills, where
field regularity varies sharply across the domain, leading to systematic
miscalibration of the predictive variance~\cite{yanes2024deep}.

Deep generative models offer a flexible alternative capable of capturing
non-stationary spatial structure. Variational Autoencoders trained on
physics-based simulators have demonstrated strong reconstruction performance for
oil spill scenarios~\cite{casado2025vae}, and convolutional encoder-decoder
architectures such as U-Net~\cite{ronneberger2015unet} generalize well across
observation modalities at inference speeds compatible with onboard deployment.
However, standard deterministic architectures produce point estimates with no
notion of predictive confidence, preventing their direct integration into active
sensing pipelines. Several approaches address this: Monte Carlo
Dropout~\cite{gal2016dropout} interprets dropout retained at inference as
approximate Bayesian inference; Deep Ensembles~\cite{lakshminarayanan2017simple}
derive epistemic uncertainty from inter-member disagreement; and Evidential Deep
Learning~\cite{amini2020deep} places a Normal-Inverse-Gamma prior over the
likelihood, enabling closed-form uncertainty decomposition from a single forward
pass. A recent comparative study showed that Deep Ensembles achieve the best
trade-off between calibration quality and computational cost under the
heterogeneous observation models characteristic of surface and aerial monitoring
platforms~\cite{yanes2026uncertainty}.

The problem of directing a sensing platform toward maximally informative
locations is most naturally formulated within Bayesian
optimization~\cite{shahriari2016bayesian}, and translates to Informative Path
Planning in the spatial field reconstruction
setting~\cite{chen2019robotic,yanes2023censored}. A key structural property of
these objectives is submodularity: greedy maximization is guaranteed to achieve
at least a $(1 - 1/e)$ fraction of the optimal
solution~\cite{nemhauser1978analysis}, justifying one-step-lookahead planners
despite their myopic nature. More expressive planners have also been explored,
including Monte Carlo Tree Search~\cite{browne2012mcts} and orienteering
formulations solved with metaheuristics~\cite{vansteenwegen2011orienteering},
which plan paths that collectively maximize an information criterion under a
budget constraint.

\section{Methods}\label{sec:methods}

To obtain a calibrated-uncertainty reconstruction model, we employ a Deep Ensemble of $M$ U-Net members trained on physics-based oil spill simulations,
as is described in \cite{yanes2026uncertainty}.
All ensemble members share a common U-Net backbone~\cite{ronneberger2015unet}
that takes as input two spatial matrices of size $H \times W$: an observation
mask $\mathbf{M} \in \{0,1\}^{H \times W}$, indicating which locations have
been visited, and a value matrix $\mathbf{V} \in \mathbb{R}^{H \times W}$,
containing the corresponding sensor readings at visited cells and zero
elsewhere. These are concatenated along the channel dimension to form
$\mathbf{X} = [\mathbf{M}, \mathbf{V}] \in \mathbb{R}^{2 \times H \times W}$.
The encoder comprises four blocks of two convolutional layers with batch
normalization and ReLU activations followed by max-pooling, with channel
progression $32 \to 512$ and a bottleneck of $1024$ channels. The decoder
mirrors this structure via transposed convolutions with skip connections,
and each member produces two output heads via a final $1 \times 1$
convolution: a mean estimate $\hat{f}^{(m)}(\mathbf{x})$ and a log-variance
$\log \hat{\sigma}^{2(m)}_{\text{ale}}(\mathbf{x})$ representing aleatoric
uncertainty. Each member is trained independently from a different random
initialization by minimizing the heteroscedastic negative log-likelihood:

\begin{equation}
\mathcal{L}_{\text{NLL}} = \frac{1}{|\mathcal{X}|}
\sum_{\mathbf{x} \in \mathcal{X}} \left[
\frac{\left(f(\mathbf{x}) - \hat{f}^{(m)}(\mathbf{x})\right)^2}
{2\hat{\sigma}^{2(m)}_{\text{ale}}(\mathbf{x})}
+ \frac{1}{2} \log \hat{\sigma}^{2(m)}_{\text{ale}}(\mathbf{x})
\right]
\label{eq:nll}
\end{equation}

At inference time, the ensemble of $M$ members produces a predictive mean
and a decomposition of total uncertainty into epistemic and aleatoric
components~\cite{lakshminarayanan2017simple,yanes2026uncertainty}:

\begin{equation}
\hat{f}(\mathbf{x}) = \frac{1}{M} \sum_{m=1}^{M} \hat{f}^{(m)}(\mathbf{x}),
\label{eq:ensemble_mean}
\end{equation}

\begin{equation}
\begin{aligned}
\hat{\sigma}^2_{\text{epi}}(\mathbf{x}) &= \frac{1}{M}
\sum_{m=1}^{M} \left(\hat{f}^{(m)}(\mathbf{x}) - \hat{f}(\mathbf{x})\right)^2, \\
\hat{\sigma}^2_{\text{ale}}(\mathbf{x}) &= \frac{1}{M}
\sum_{m=1}^{M} \hat{\sigma}^{2(m)}_{\text{ale}}(\mathbf{x})
\end{aligned}
\label{eq:ensemble_unc}
\end{equation}

Epistemic uncertainty $\hat{\sigma}^2_{\text{epi}}$ reflects inter-member
disagreement and is reducible by collecting additional observations, making
it the actionable component for path planning. Aleatoric uncertainty
$\hat{\sigma}^2_{\text{ale}}$ captures irreducible sensor noise and remains
invariant to further sampling. The ensemble is trained offline on a dataset
of physics-based oil spill simulations and evaluated at inference without
retraining, yielding reconstruction and uncertainty maps in a single forward
pass through all members.

The resulting Ensemble model is calibrated in the sense that the epistemic uncertainty map $\hat{\sigma}^2_{\text{epi}}$ is well-aligned with the true reconstruction error, as demonstrated in~\cite{yanes2026uncertainty}. 
This property is crucial for Informative Path Planning, as it ensures that the vehicle can reliably 
identify regions where additional measurements will yield the greatest improvement in field reconstruction.

\subsection{Gaussian Process Baseline}

As a probabilistic baseline, we employ a Gaussian Process~\cite{rasmussen2006gaussian}
with a composite kernel combining a squared exponential term with an additive
white noise component:

\begin{equation}
k(\mathbf{x}, \mathbf{x}') = \sigma^2_f
\exp\!\left(-\frac{\|\mathbf{x} - \mathbf{x}'\|^2}{2\ell^2}\right)
+ \sigma^2_n \, \delta(\mathbf{x}, \mathbf{x}')
\label{eq:gp_kernel}
\end{equation}

where $\sigma^2_f$ is the signal variance, $\ell$ is the length-scale, and
$\sigma^2_n$ is the noise variance. The three hyperparameters
$\{\sigma^2_f, \ell, \sigma^2_n\}$ are optimized per test sample by maximizing
the marginal log-likelihood via L-BFGS. The GP posterior mean provides the
field reconstruction, while the posterior variance serves as the uncertainty
estimate for planning. Crucially, the white noise term $\sigma^2_n$ is a
global scalar that imposes a spatially uniform aleatoric component regardless
of local observation density or field regularity. Combined with the isotropic
RBF kernel, which enforces a single global length-scale across the entire
domain, the GP is fundamentally misspecified for oil spill fields, where sharp
gradients at contamination boundaries coexist with large smooth background
regions. This mismatch produces miscalibrated uncertainty estimates that
degrade the reliability of the GP as an active sensing criterion, as
demonstrated in~\cite{yanes2026uncertainty}.

\subsection{Path Planning Algorithms}
\label{sec:planners}

All planners share a common reactive planning loop regardless of their internal
strategy. At each decision step $t$, the vehicle receives the current uncertainty
map $\hat{\sigma}^2_{\text{epi}}(\cdot \mid \mathcal{D}_t)$ produced by the
ensemble conditioned on all observations collected so far. The planner is then
invoked with the remaining budget $T - t$ and produces a complete candidate
trajectory. Only the first action of that trajectory is executed: the vehicle
moves to the selected waypoint, collects a new observation, and incorporates
it into $\mathcal{D}_{t+1}$. The ensemble is queried again to produce an updated
reconstruction and uncertainty map, and the planning cycle restarts. This
receding-horizon scheme ensures that every planning decision is grounded in the
most recent observations, so that the trajectory adapts continuously to the
information revealed during the mission rather than committing to a plan computed
from the initial state.

\subsubsection{Greedy Planners}

Greedy strategies select the next waypoint $\mathbf{x}_{t+1}$ by maximizing
a scalar criterion over the set of unvisited reachable locations
$\mathcal{X} \setminus \mathcal{D}_t$ without lookahead:

\begin{equation}
\mathbf{x}_{t+1} = \arg\max_{\mathbf{x} \in \mathcal{X} \setminus \mathcal{D}_t}
\; a(\mathbf{x}),
\label{eq:greedy}
\end{equation}

where $a(\mathbf{x})$ is the acquisition function. We consider two variants.
The uncertainty-greedy planner sets $a(\mathbf{x}) = \hat{\sigma}^2_{\text{epi}}(\mathbf{x})$,
directing the vehicle toward the location of maximum epistemic uncertainty.
The model-greedy planner sets $a(\mathbf{x}) = \hat{f}(\mathbf{x})$, directing
the vehicle toward the location of maximum predicted field value, which is
appropriate when the monitoring objective is to locate and characterize the
peak of the contaminant distribution rather than to reduce global reconstruction
error. Both greedy variants are myopic but computationally negligible, and the
uncertainty-greedy criterion is theoretically motivated by the submodularity
of the information gain objective~\cite{nemhauser1978analysis}: each observation
reduces the epistemic uncertainty that subsequent measurements can eliminate,
so that greedily targeting the maximum residual uncertainty achieves a
$(1 - 1/e)$ approximation of the optimal multi-step policy.

\subsubsection{Monte Carlo Tree Search}

MCTS~\cite{browne2012mcts} extends the greedy approach by performing a
multi-step lookahead through a stochastic tree expansion. At each planning
invocation, MCTS builds a search tree rooted at the current vehicle state
$(\mathbf{x}_t, \mathcal{D}_t)$ and alternates between four standard phases:
selection via the UCT criterion~\cite{kocsis2006bandit}, expansion, rollout
using the uncertainty-greedy heuristic, and backpropagation of the cumulative
epistemic uncertainty along the simulated path.

A key design contribution is an uncertainty-adaptive branching strategy: at
each node corresponding to location $\mathbf{x}$, the neighborhood radius
$r(\mathbf{x})$ from which candidate children are sampled is inversely
proportional to the local epistemic uncertainty,

\begin{equation}
r(\mathbf{x}) = \max\left(r_{\min}, \min \left[\frac{r_{\max} - r_{\min}}
{\hat{\sigma}_{\text{epi}}(\mathbf{x}) + \epsilon} + r_{\min}, r_{\max}
\right]\right)
\label{eq:adaptive_radius}
\end{equation}

where $\epsilon > 0$ ensures numerical stability and $r_{\min}, r_{\max}$ 
are predefined radius bounds. This focuses the search
budget on regions of high uncertainty, where careful multi-step planning yields
the greatest benefit, while pruning branches in well-characterized areas.
After a fixed number of iterations, the vehicle commits to the first action
of the highest-value branch, consistent with the receding-horizon loop.

\subsubsection{Orienteering with Simulated Annealing}

The orienteering problem formulation treats path planning as a combinatorial
optimization problem~\cite{vansteenwegen2011orienteering}: candidate waypoints
are sampled on a regular sub-grid of the map, and the planner selects a subset
and an ordering that maximizes total reward while respecting a budget constraint
on trajectory cost. The reward of traversing an edge is defined as the sum of
epistemic uncertainty values over the pixels of the connecting segment, and the
cost as the Euclidean distance between waypoints. This allows the planner to
reason about the collective value of a sequence of waypoints rather than
committing greedily at each step.

The orienteering problem is solved with a two-phase metaheuristic. First, a
multi-start greedy construction generates a diverse set of initial solutions by
building paths greedily according to reward-to-cost efficiency under different
random orderings of the candidate list. Second, a hill-climbing local search
refines the best solution by alternating between removal of interior waypoints
that increase total reward and insertion of unvisited candidates at positions
that increase reward without exceeding the budget, until no single move yields
further improvement.
\section{Experimental Results}
\label{sec:results}

\subsection{Experimental Setup}

\subsubsection{Simulation Environment}

Experiments are conducted on physics-based oil spill scenarios generated by a
particle simulator on a $100 \times 100$ grid. Spills are initialised at random
locations and evolved under stochastic wind and tidal dynamics, producing
realizations that vary widely in shape, extent, and gradient structure. All
fields are normalised to $[0, 1]$, and a contact point sensor returns field
values with additive Gaussian noise $\varepsilon_i \sim \mathcal{N}(0, 0.1^2)$.

The uncertainty model follows~\cite{yanes2026uncertainty}: five independent
U-Net members with channel progression $32 \to 512$ and a bottleneck of $1024$
channels, each producing a mean and log-variance map via a heteroscedastic
output head trained offline by minimising Eq.~(\ref{eq:nll}). At inference, the
epistemic uncertainty map is computed as inter-member variance
(Eq.~\ref{eq:ensemble_unc}) and consumed by all planners without further
adaptation.

Five policies are evaluated: $\epsilon$-Greedy (random exploration
decaying from $\varepsilon{=}0.10$ to $0.01$); Value Greedy
($a(\mathbf{x}) = \hat{f}(\mathbf{x})$); Uncertainty Greedy
($a(\mathbf{x}) = \hat{\sigma}^2_{\text{epi}}(\mathbf{x})$); Receding
Horizon Orienteering (planning horizon $30$, actuation horizon $5$, grid
resolution $2$ cells); and MCTS ($2000$ simulations, depth equal to
remaining budget, $c{=}1.414$, $\gamma{=}0.8$, radii $r \in [1,4]$, tree
reuse enabled). All combinations are evaluated over a common held-out test
partition following the receding-horizon loop of Section~\ref{sec:planners}.

All experiments and code is available at \url{https://gitlab.ratatosk.cc/syanes/uncertainty-online-planning-caepia-2026}.
\subsection{Evaluation Metrics}

Reconstruction accuracy is measured by the Root Mean Squared Error between the
ensemble mean $\hat{f}$ and the ground-truth field $f$ over all unvisited cells:

\begin{equation}
\mathrm{RMSE} = \sqrt{\frac{1}{|\mathcal{X}|}
\sum_{\mathbf{x} \in \mathcal{X}} \left(f(\mathbf{x}) -
\hat{f}(\mathbf{x})\right)^2}.
\label{eq:rmse}
\end{equation}

To account for the varying difficulty across episodes, RMSE is normalised by the
value recorded at the first decision step, so the metric reflects relative
improvement rather than absolute map difficulty. We additionally report the
Intersection over Union (IoU)~\cite{casado2025vae} between the predicted and
ground-truth contamination masks, obtained by thresholding both fields at a fixed
value $\tau$:

\begin{equation}
\mathrm{IoU} = \frac{|\hat{\mathcal{S}} \cap \mathcal{S}|}{|\hat{\mathcal{S}}
\cup \mathcal{S}|},
\label{eq:iou}
\end{equation}

where $\mathcal{S} = \{\mathbf{x} : f(\mathbf{x}) > \tau\}$ and
$\hat{\mathcal{S}} = \{\mathbf{x} : \hat{f}(\mathbf{x}) > \tau\}$. IoU
captures the spatial precision of spill localisation independently of global
reconstruction error.

\subsection{Results}

\subsubsection{Model Comparison and Policy–Model Interaction}

\begin{figure}[t]
\centering
\includegraphics[width=\linewidth]{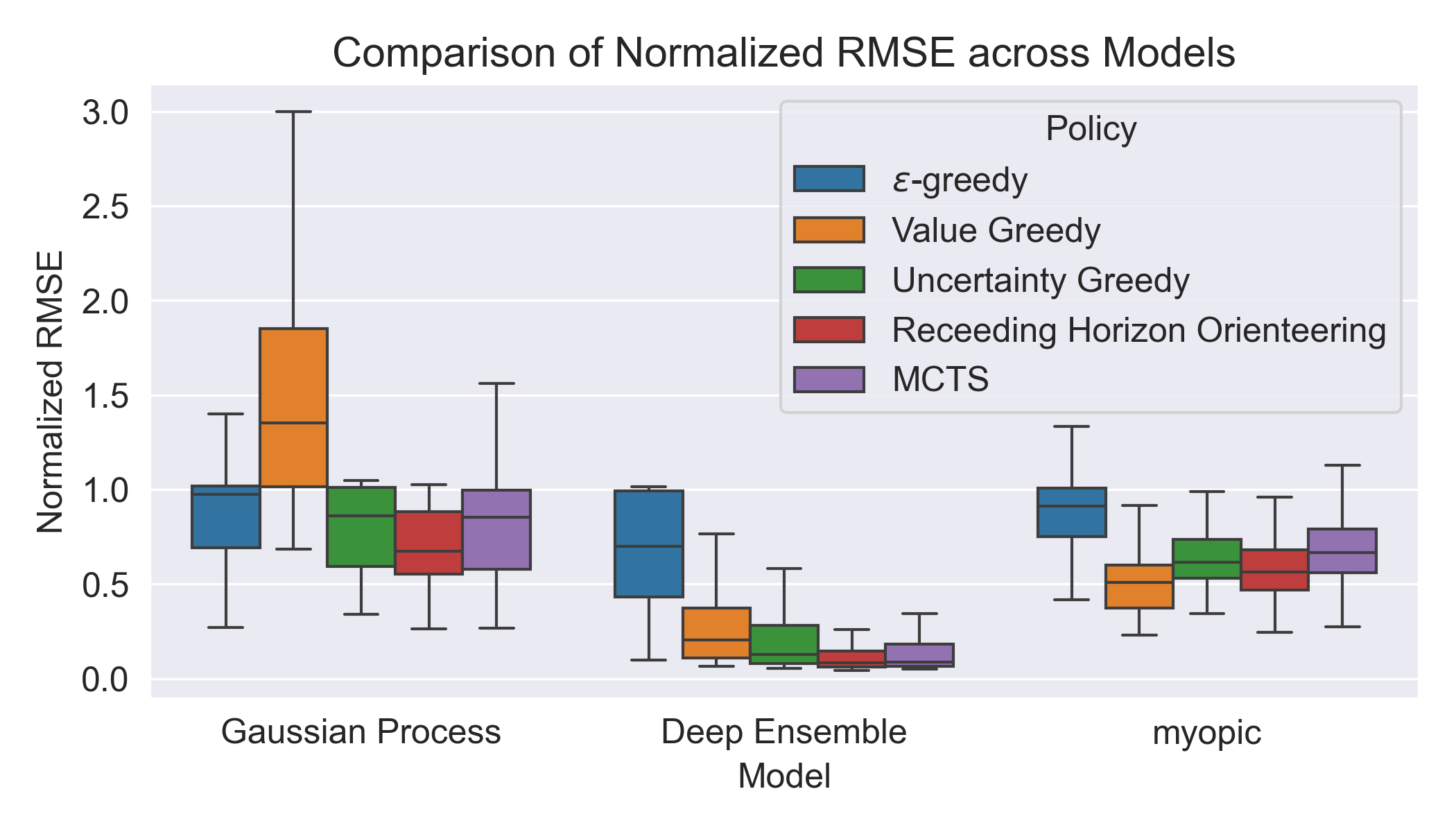}
\caption{Distribution of normalised RMSE for each (model, policy) combination
across all evaluation episodes. Each group of five boxes corresponds to one
reconstruction model; colours denote policies. Lower is better.}
\label{fig:boxplot_models}
\end{figure}

Table~\ref{tab:model_rmse} and Figure~\ref{fig:boxplot_models} reveal two
complementary findings. The choice of reconstruction model is the dominant
factor in planning performance: the Deep Ensemble achieves a mean normalised
RMSE of $0.1568$, compared to $0.6243$ for the myopic baseline and $0.9350$
for the Gaussian Process, representing reductions of $75\%$ and $83\%$
respectively. Moreover, the quality of the uncertainty signal determines how
much the policy choice matters: the inter-policy spread is negligible under a
miscalibrated model, but widens substantially when uncertainty aligns with the
true reconstruction error.

The myopic baseline reconstructs the field via $k$-Nearest Neighbours
interpolation and models uncertainty through inverse-distance weighting,
producing values in $[0,1]$ that reflect spatial coverage rather than
predictive confidence. Under this model, as under the Gaussian Process, all
policies converge to a narrow and uniformly mediocre band: without a principled
uncertainty signal, the reconstruction ceiling is set by the model rather than
the planner, and sophisticated multi-step algorithms offer no advantage over
random exploration. The GP additionally suffers from high variance
(std $= 0.5618$), confirming that its stationary isotropic kernel is
fundamentally misspecified for this kind of fields~\cite{yanes2026uncertainty}.

\begin{table}[t]
\centering
\caption{Normalised RMSE per model, averaged over all policies and episodes
(mean $\pm$ std). Best result in \textbf{bold}.}
\label{tab:model_rmse}
\begin{tabularx}{\linewidth}{Xc}
\toprule
Model & RMSE (norm.) \\
\midrule
Gaussian Process & $0.9350 \pm 0.5623$ \\
Myopic           & $0.6243 \pm 0.2300$ \\
Deep Ensemble    & $\mathbf{0.1568} \pm 0.2856$ \\
\bottomrule
\end{tabularx}
\end{table}

With the Deep Ensemble, the overall error collapses and the performance differences 
between policies become more pronounced, revealing a clear hierarchy among planning 
strategies that was invisible under weaker models. This amplification effect 
is the central empirical finding of this paper: a well-calibrated uncertainty 
signal transforms the uncertainty map into a reliable planning gradient that differentiates
effective exploration strategies from ineffective ones.

\subsubsection{Policy Comparison}

\begin{figure*}[t]
    \centering
    \subfloat[Normalised RMSE over time.]{\includegraphics[width=0.49\linewidth]{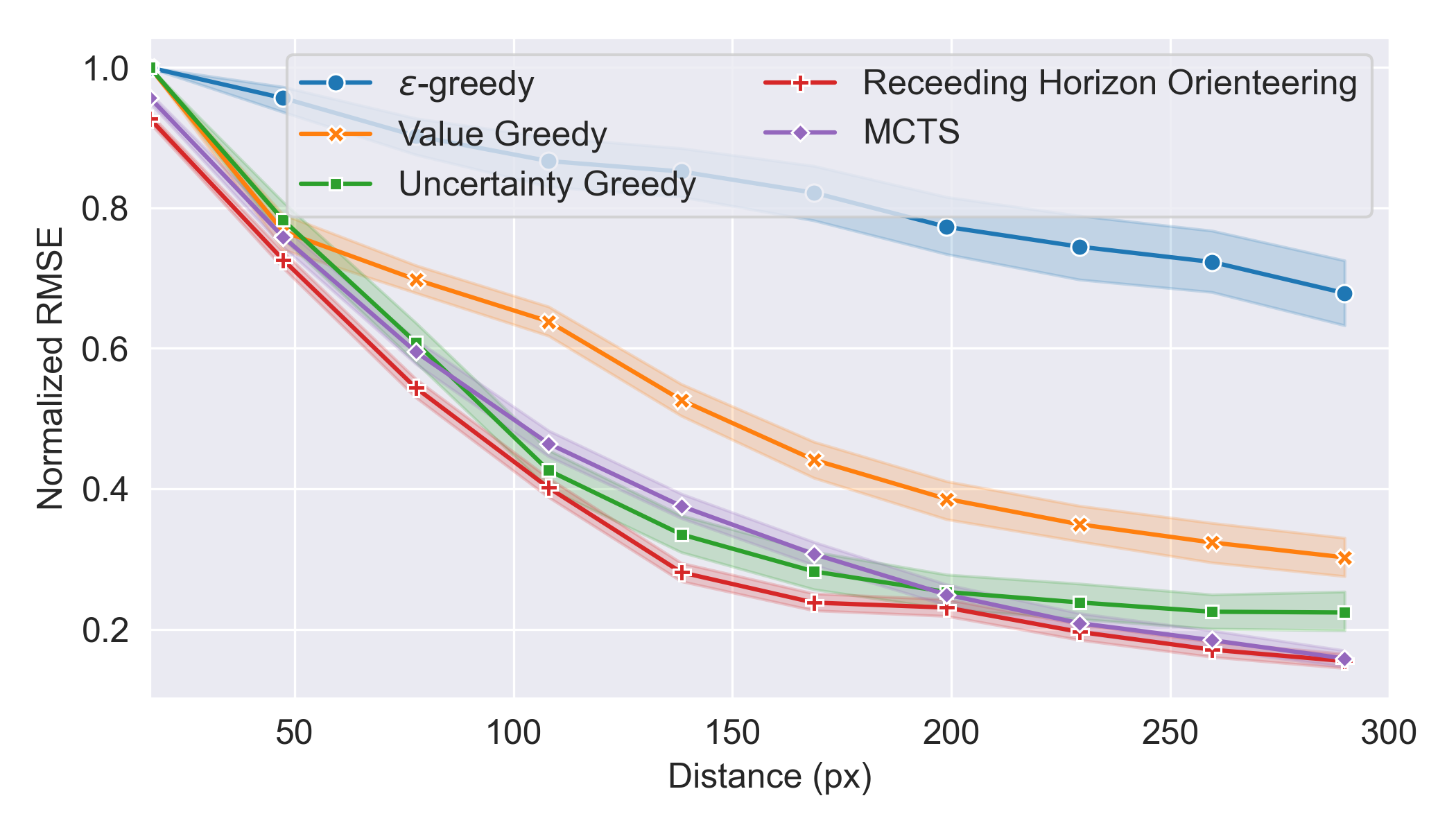}\label{fig:rmse_time}}
    \hfill
    \subfloat[IoU over time.]{\includegraphics[width=0.49\linewidth]{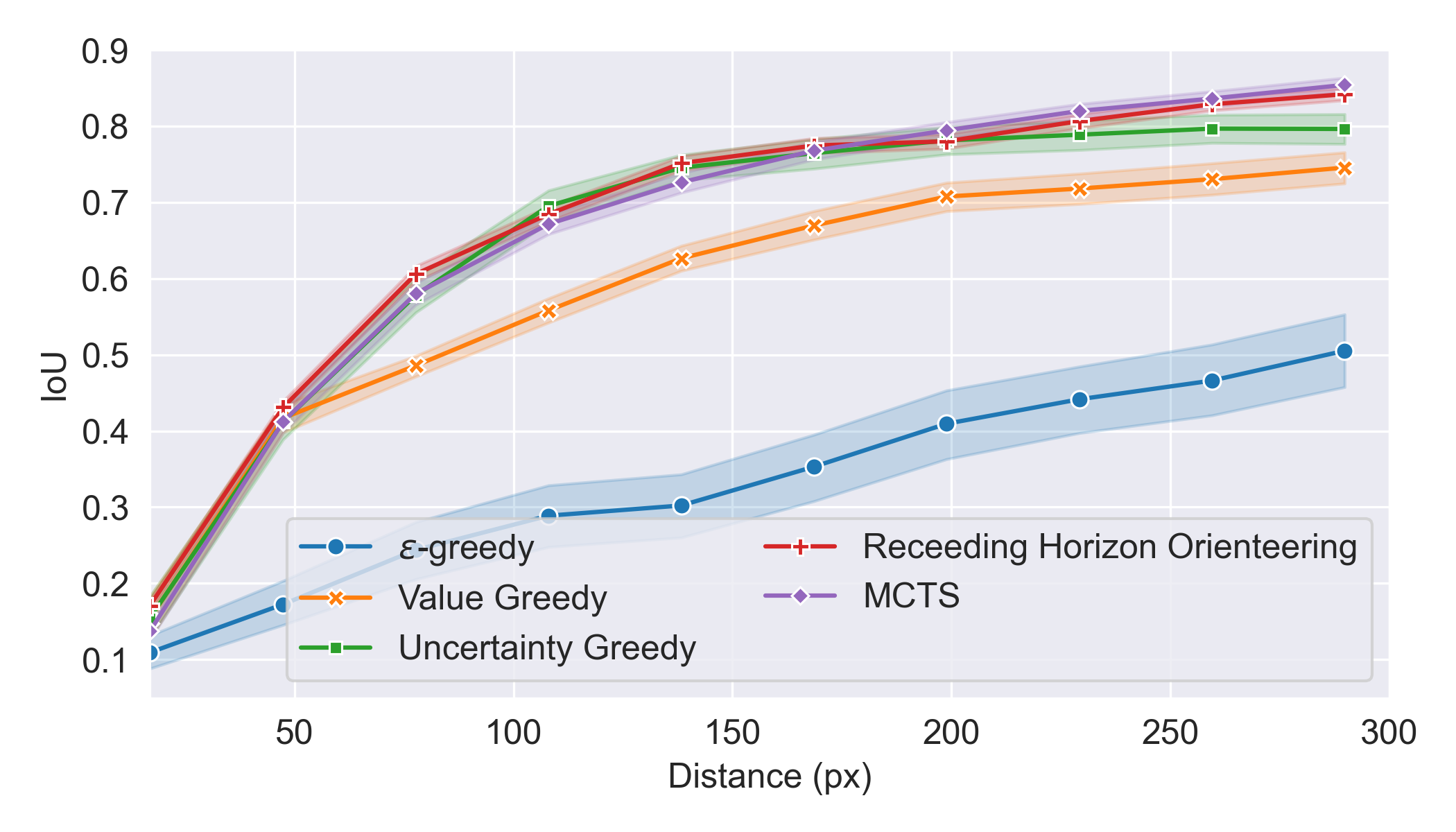}\label{fig:iou_time}}
    \caption{Evolution of reconstruction metrics across decision steps for each
    policy under the Deep Ensemble model. Shaded bands denote $\pm 1$ standard
    deviation over all evaluation episodes.}
    \label{fig:metrics_over_time}
\end{figure*}

Table~\ref{tab:policy_rmse} reports the normalised RMSE and IoU for each policy
under the Deep Ensemble model. In Fig. \ref{fig:metrics_over_time}, the evolution over
time is illustrated for each policy using the Deep Ensemble model. 
The Receding Horizon Orienteering planner achieves the lowest reconstruction error ($0.1434 \pm 0.1385$), followed closely by MCTS
($0.1561 \pm 0.1473$); both outperform the greedy Uncertainty criterion
($0.2097 \pm 0.1823$) by approximately $32\%$ and $26\%$ respectively. The
ranking is reversed for localisation: MCTS attains the highest IoU
($0.8563 \pm 0.1099$), marginally ahead of Orienteering ($0.8491 \pm 0.1122$)
and well above Uncertainty Greedy ($0.8085 \pm 0.1315$). Taken together, these
results show that, given a well-calibrated epistemic map, multi-step lookahead
\emph{does} pay off: both MCTS and Orienteering consistently outperform one-step
greedy selection, with Orienteering favouring boundary coverage and global
reconstruction accuracy while MCTS yields a slight advantage in spill
localisation through its deeper tree search over candidate waypoints.

Uncertainty Greedy still substantially outperforms the non-uncertainty baselines,
confirming the primacy of the epistemic signal over the planning horizon.
Value Greedy achieves moderate results ($\mathrm{RMSE} = 0.2794$,
$\mathrm{IoU} = 0.7594$): concentrating observations near the predicted
contamination peak provides a useful proxy for localisation but neglects
uncertain boundary regions, limiting global reconstruction quality. The
$\epsilon$-Greedy baseline performs worst on both metrics, confirming that undirected
exploration is an ineffective use of the measurement budget regardless of the
quality of the underlying model.

\begin{table}[t]
\centering
\caption{Normalised RMSE and IoU for each policy using the Deep Ensemble model
(mean $\pm$ std). Best result per metric in \textbf{bold}.}
\label{tab:policy_rmse}
\footnotesize
\setlength{\tabcolsep}{3pt}
\begin{tabularx}{\linewidth}{Xcc}
\toprule
Policy & RMSE (norm.) & IoU \\
\midrule
$\epsilon$-Greedy             & $0.6774 \pm 0.3024$ & $0.5053 \pm 0.3043$ \\
Value Greedy                  & $0.2794 \pm 0.2183$ & $0.7594 \pm 0.1621$ \\
Uncertainty Greedy            & $0.2097 \pm 0.1823$ & $0.8085 \pm 0.1315$ \\
Receding Horizon Orienteering & $\mathbf{0.1434} \pm 0.1385$ & $0.8491 \pm 0.1122$ \\
MCTS                          & $0.1561 \pm 0.1473$ & $\mathbf{0.8563} \pm 0.1099$ \\
\bottomrule
\end{tabularx}
\end{table}

\subsubsection{Qualitative Analysis}

\begin{figure*}[t]
\centering
\includegraphics[width=\textwidth]{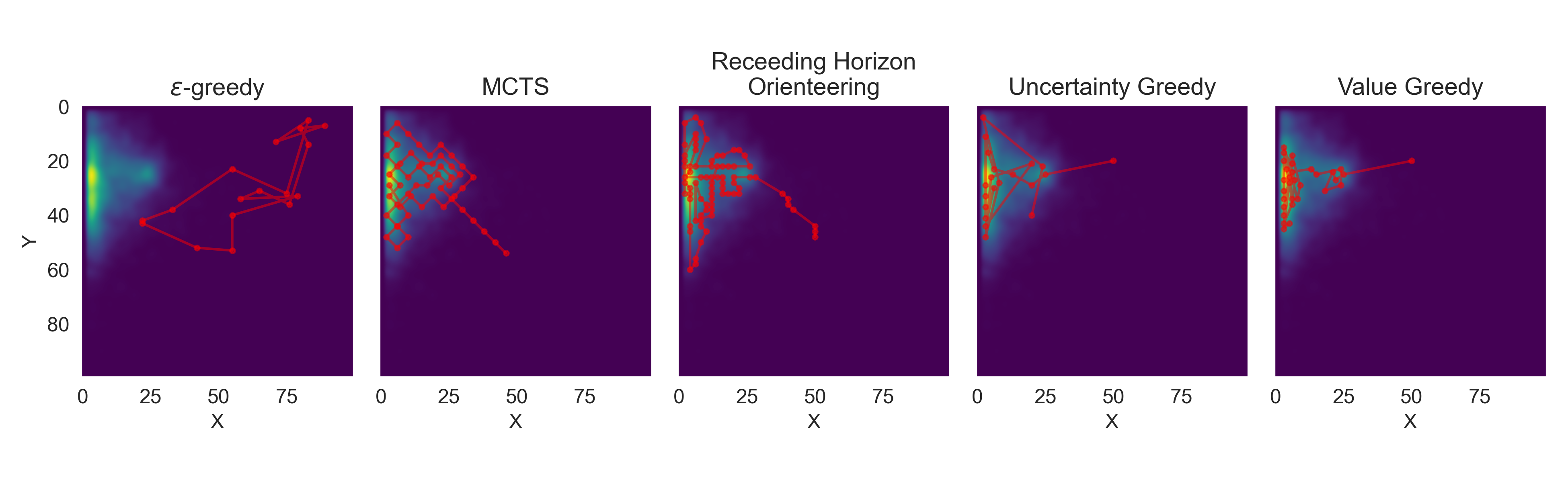}
\caption{Example trajectories for each policy on a representative scenario
(Deep Ensemble model). The background shows the ground-truth oil spill field;
red dots and lines mark visited waypoints and the connecting path.
Multi-step planners (MCTS, Orienteering) produce structured boundary coverage,
whereas $\epsilon$-Greedy disperses waypoints across uninformative background
regions.}
\label{fig:trajectories}
\end{figure*}

Figure~\ref{fig:trajectories} illustrates the trajectories produced by each
policy for a representative scenario. The contaminated region is concentrated
in the upper-left quadrant, where ensemble uncertainty is highest at mission
start. The $\epsilon$-Greedy planner disperses waypoints across the full domain,
wasting the majority of the budget on uninformative background cells. Value
Greedy concentrates near the spill peak but revisits already-characterised areas
rather than systematically covering the boundary. By contrast, Uncertainty
Greedy, MCTS, and Receding Horizon Orienteering all converge to dense, structured
coverage of the spill boundary, where residual epistemic uncertainty is highest
and additional observations yield the greatest marginal reduction in
reconstruction error. MCTS produces a systematic grid-like sweep of the
uncertain region, a consequence of its tree search explicitly reasoning over
multi-step action sequences, while the orienteering planner generates compact
sub-paths that collectively maximise accumulated uncertainty reward under the
distance budget constraint, covering slightly more boundary area at the cost
of a marginally lower IoU.

\section{Conclusion}
\label{sec:conclusion}

This paper has presented a controlled evaluation of five path planning strategies
for scalar field reconstruction under a well-calibrated deep uncertainty model,
with the explicit goal of isolating the contribution of the planning algorithm
from that of the uncertainty model. The central finding is that model quality
is the dominant factor in path planning performance: replacing a stationary
Gaussian Process with a Deep Ensemble reduces the normalised RMSE by $83\%$
across all policies, and the myopic baseline by $75\%$. Crucially, this
improvement is not uniform across planners. A well-calibrated uncertainty map
amplifies the importance of the planning strategy, revealing a clear performance
hierarchy that is invisible under miscalibrated models.

Among the planning algorithms evaluated, multi-step lookahead methods
consistently outperform one-step greedy selection when the uncertainty signal
is reliable. Receding Horizon Orienteering achieves the lowest reconstruction
error ($0.1434$), while MCTS attains the highest spill localisation IoU
($0.8563$), both substantially ahead of the Uncertainty Greedy baseline
($0.2097$, $0.8085$). These results confirm that the theoretical
$(1-1/e)$ approximation guarantee of greedy submodular maximisation, while
non-trivial, leaves meaningful room for improvement when a richer planning
horizon is available.

From a practical deployment perspective, MCTS emerges as a particularly
attractive option. Although Receding Horizon Orienteering marginally outperforms
it on RMSE, the metaheuristic solver, which combines multi-start greedy construction
with hill-climbing local search, incurs a computational cost that is an order
of magnitude higher than MCTS per planning call. MCTS, by contrast, amortises
its $2000$ simulations efficiently through tree reuse across receding-horizon
cycles and its uncertainty-adaptive branching strategy, which concentrates the
search budget precisely where the epistemic signal is most informative. For
resource-constrained platforms such as ASVs operating
under real-time replanning requirements, this cost differential is decisive.
MCTS offers near-optimal reconstruction and superior localisation at a fraction
of the computational overhead, making it the recommended planner when both
performance and efficiency are required.

Future work will extend this framework in two directions. First, the online
integration of Evidential Deep Learning, which achieves comparable calibration
to the ensemble at a single forward pass, could further reduce inference
latency without sacrificing uncertainty quality. Second, extending the evaluation
to multi-agent settings, where the epistemic uncertainty map must be shared and
updated across vehicles, will require coordination-aware planning algorithms
that remain efficient under communication constraints.

\section*{Acknowledgment}
Proyecto PID2024-158365OB-C21 financiado por
MICIU/AEI/10.13039/501100011033 y por FEDER, UE.

\section*{Competing Interests}
The authors have no competing interests to declare that are
relevant to the content of this article.

\bibliographystyle{IEEEtran}
\bibliography{biblio}

\end{document}